\documentclass[11pt]{article}

\usepackage[final]{acl}

\usepackage{times}
\usepackage{latexsym}

\usepackage[T1]{fontenc}

\usepackage[utf8]{inputenc}

\usepackage{microtype}

\usepackage{inconsolata}

\usepackage{graphicx}

\usepackage{amssymb}
\usepackage{amsmath}
\usepackage{booktabs}
\usepackage{multirow}
\usepackage{enumitem}

\title{RAG-KT: Cross-platform Explainable Knowledge Tracing with \\ Multi-view Fusion Retrieval Generation}

\author{
 \textbf{Zhiyi Duan\textsuperscript{1}},
 \textbf{Hongyu Yuan\textsuperscript{1}},
 \textbf{Rui Liu\textsuperscript{1}\thanks{Corresponding author}},
\\
 \textsuperscript{1}Inner Mongolia University, Hohhot, China,\\
   \texttt{duanzy@imu.edu.cn}, \texttt{22509014@mail.imu.edu.cn}, \texttt{imucslr@imu.edu.cn} \\
\\
}

\begin{document}
\maketitle
\begin{abstract}
Knowledge Tracing (KT) infers a student’s knowledge state from past interactions to predict future performance. 
Conventional Deep Learning (DL)-based KT models are typically tied to platform-specific identifiers and latent representations, making them hard to transfer and interpret.
Large Language Model (LLM)-based methods can be either ungrounded under prompting or overly domain-dependent under fine-tuning.
In addition, most existing KT methods are developed and evaluated under a same-distribution assumption. 
In real deployments, educational data often arise from heterogeneous platforms with substantial distribution shift, which often degrades generalization. 
To this end, we propose \textbf{RAG-KT}, a retrieval-augmented paradigm that frames cross-platform KT as reliable context constrained inference with LLMs. 
It builds a unified multi-source structured context with cross-source alignment via \textbf{Question Group} abstractions and retrieves complementary rich and reliable context for each prediction, enabling grounded prediction and interpretable diagnosis. 
Experiments on three public KT benchmarks demonstrate consistent gains in accuracy and robustness, including strong performance under cross-platform conditions.
\end{abstract}

\section{Introduction}

\begin{figure}[!t]
    \centering
    \includegraphics[width=1\linewidth]{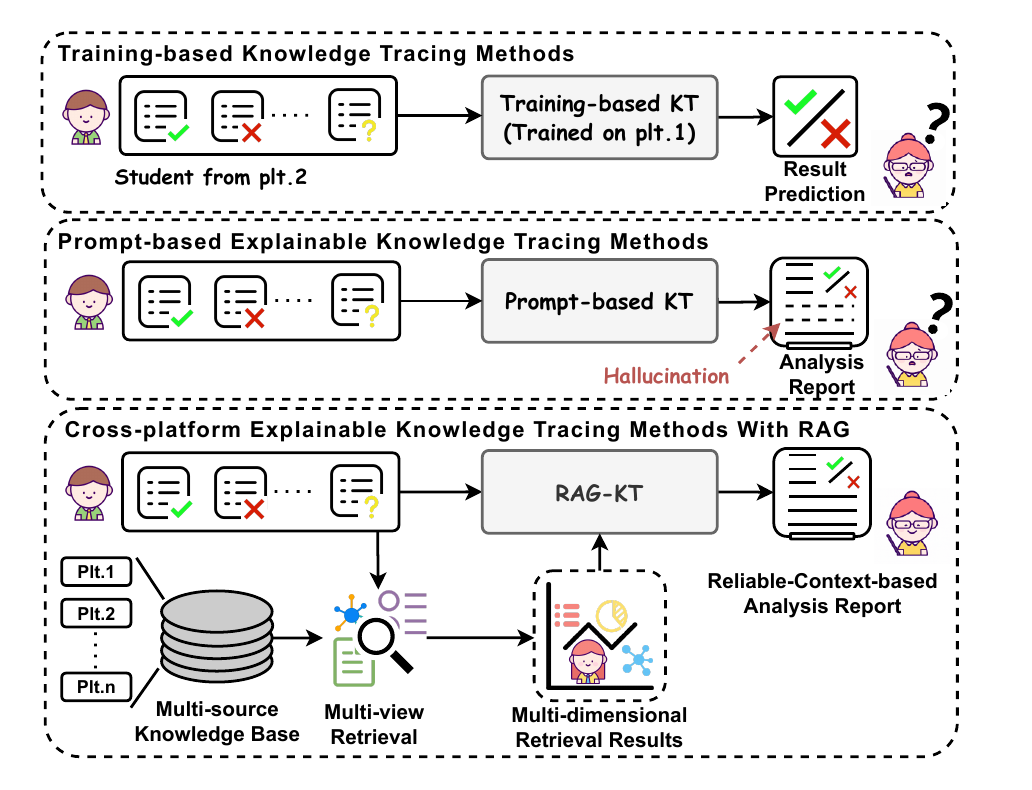}
    \caption{Comparison of KT paradigms. Traditional models struggle with cross-platform shifts, while prompt-based LLMs are prone to hallucinations. RAG-KT overcomes these by leveraging multi-view retrieval to enable reliable, grounded inference.}
    \label{fig:fig1}
\end{figure}
Knowledge Tracing (KT) is a fundamental task in educational data mining that models a student’s knowledge state from historical responses to predict future performance \cite{bkt}. 
As a cornerstone of personalized learning, KT has been extensively studied and has made notable progress \cite{ktsurvey}.
Despite these advances, a critical gap remains between common KT research settings and real-world deployment.

Conventional Deep Learning (DL)-based KT approaches often rely on closed, pre-defined identifiers, which limits their ability to generalize to unseen questions, new knowledge concepts, or different platforms \cite{cold-start}. 
Moreover, these models typically encode knowledge states as latent vectors, which hinders interpretability and constrains their usefulness for pedagogical decision-making \cite{ding2019deep, ikt}.

Recently, Large Language Models (LLMs) have opened new opportunities for improving interpretability and flexibility \cite{gpt, qwen, deepseek, hao2025mimo}. 
With their natural language reasoning capability, LLMs can extend KT beyond numeric prediction to interpretable diagnosis by generating human-readable analytical reports \cite{edusurvey, cikt}. 
However, directly applying LLMs to KT faces a dilemma. 
Prompt-based methods are flexible, yet they often lack platform-specific learning signals and can therefore hallucinate or produce generic explanations that are not grounded in an individual’s learning history \cite{efkt, hisekt}. 
In contrast, fine-tuning LLMs on interaction data can improve in-domain performance, but it typically strengthens dependence on the source platform’s distribution, sacrifices cross-platform generalization, and incurs high retraining costs whenever the data source changes \cite{cikt, llamalora, 2tkt}. 

Overall, both DL-based KT and current LLM-based solutions are commonly developed and assessed under a same-distribution assumption. 
This assumption is largely inherited from the fact that most benchmark datasets and training pipelines are collected under a single-platform setting. 
In practice, however, educational data are generated across many heterogeneous platforms. 
Even when these platforms cover similar knowledge concepts, their student populations, interaction patterns, and question representations can still differ substantially, resulting in non-trivial distribution shift.

We argue that to handle the complexity of cross-platform data, KT must go beyond the single-distribution assumption. 
The key challenges lie in two aspects: (i) how to organize cross-platform data and unify heterogeneous information into a single, unified representation, and (ii) how to obtain information-rich and reliable context from it.
To this end, we propose RAG-KT, a retrieval-augmented \cite{rag, awecita} knowledge tracing paradigm.
As shown in Fig.~\ref{fig:fig1}, our proposed framework addresses the interpretability and cross-platform capability challenges by augmenting LLMs with structured retrieval and external knowledge integration.

Specifically, RAG-KT builds a unified heterogeneous graph that integrates interaction signals across platforms. 
To mitigate identifier inconsistency and enable cross-source alignment, we introduce \textbf{Question Group} nodes as an intermediate abstraction layer that aggregates questions with similar instructional attributes. 
This design allows structured reasoning even when question IDs are not shared across platforms. 
Building on this representation, we further design a multi-dimensional retrieval mechanism to collect complementary context for the current prediction step. 
The retrieved context is then organized into structured, context-grounded prompts that constrain the LLM to produce the final prediction and an interpretable analysis report.
Experimental results demonstrate that the retrieval-augmented structured context in RAG-KT consistently improves prediction accuracy and robustness. In summary, our main contributions are as follows:
\begin{itemize}
    \item We propose RAG-KT, the first retrieval-augmented paradigm for applying LLMs to KT, aiming to address the limitations of existing KT methods in interpretability and cross-platform generalizability.
    \item We construct a multi-source heterogeneous knowledge base and design a multi-view fusion retrieval framework to extract task-relevant knowledge, and generate an explainable analysis report via structured reliable-context-based prompting, thereby enhancing LLM performance on KT.
    \item We conduct comprehensive experiments on three widely used datasets, and the results show that RAG-KT achieves superior prediction accuracy, produces interpretable analytical reports, and remains robust under cross-platform scenarios.
\end{itemize}

\section{Related Work}
\subsection{Structured Knowledge Modeling in KT}
Early models use binary knowledge states and predefined skill mappings \cite{bkt}.
Subsequent neural models encode interactions as embeddings but lack semantic structure and cross-platform generalization \cite{dkt, dkvmn}.
To address this, recent works have introduced knowledge graphs and use GNNs \cite{gnn} to capture skill dependencies \cite{gkt, pebg}.
More recently, methods enhance KT by modeling temporal dynamics or phase-wise learning trajectories over evolving graph structures \cite{dygkt}.
However, these methods typically assume fixed, well-aligned datasets and do not address the challenges of integrating heterogeneous interactions from multiple platforms.
Forcing integration will only affect the quality of relation extraction, which will directly affect prediction accuracy.

\subsection{LLM-Enhanced KT Methods}

Recent studies have explored LLMs to enhance knowledge tracing, aiming to improve interpretability and address cold-start issues \cite{ragsurvey}. 
Some approaches directly fine-tune LLMs on student interaction data \cite{cikt, clst, llm-kt}. 
Meanwhile, others use few-shot or zero-shot prompting strategies \cite{efkt, lokt, hisekt}. 
Despite their promise, many existing LLM-based KT methods operate in closed or shallow contexts, lacking structured grounding from educational knowledge.
As a result, they may suffer from inconsistency or hallucinations, especially under sparse or out-of-distribution scenarios. 
In contrast, our RAG-KT framework grounds LLM inference in a structured, heterogeneous knowledge graph and retrieves personalized, multi-dimensional context. This design ensures both interpretability and generalization, especially in low-resource or unseen settings.

\section{Methodology}
\begin{figure*}[!ht]
    \centering
    \includegraphics[width=1\linewidth]{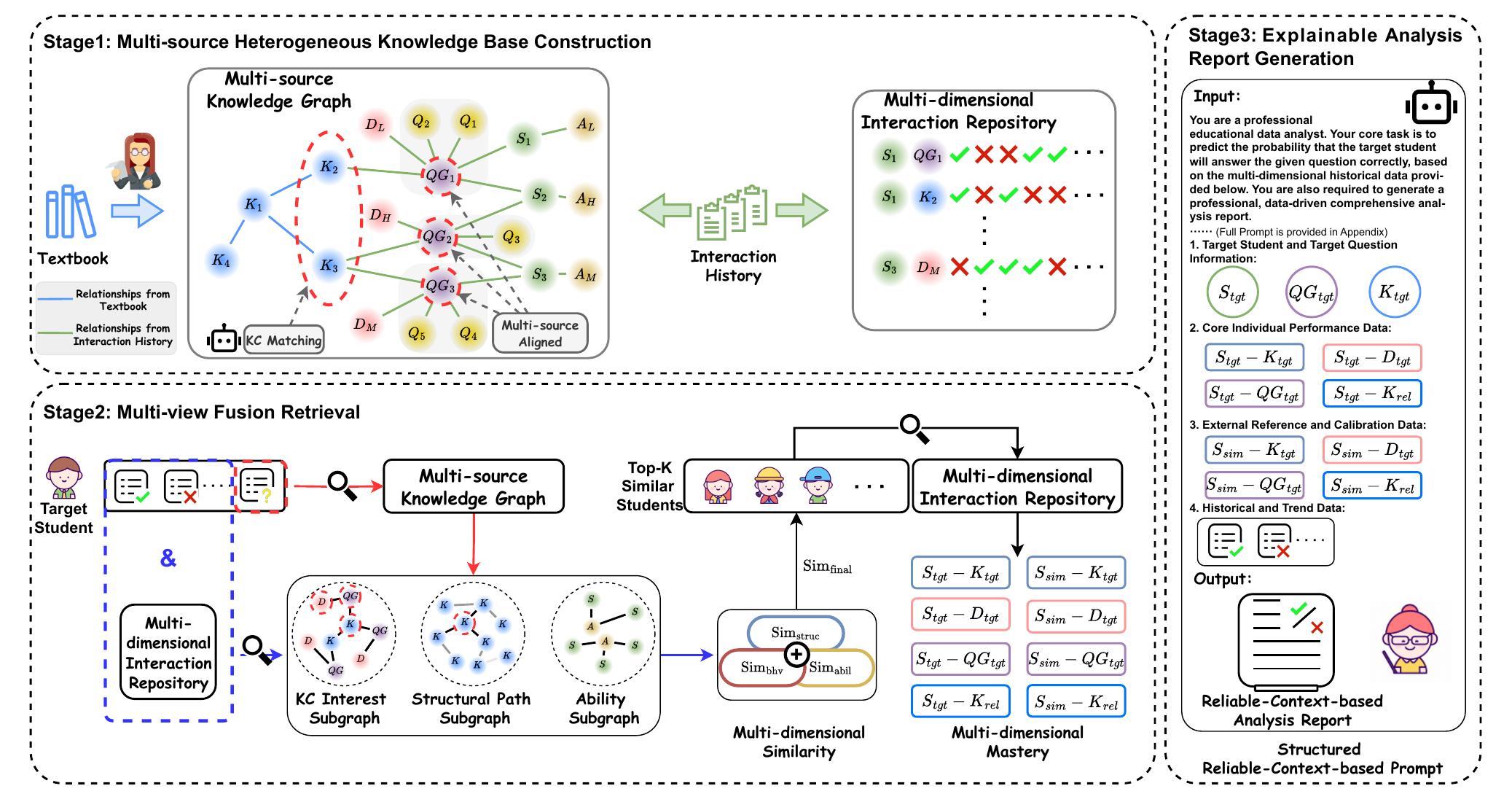}
    \caption{The RAG-KT framework, notation like $S_{tgt} – K_{tgt}$ indicates the target student's performance on the target concept, while $S_{sim} – K_{tgt}$ represents similar students’ aggregated performance on that dimension. $QG$ is a collection of semantically equivalent $Q$ from different data sources.
    }
    \label{fig:fig2}
\end{figure*}

\subsection{Overall Framework}
The overall framework of RAG-KT is shown in Fig.~\ref{fig:fig2}.
It begins by constructing a Multi-source Heterogeneous Knowledge Base $\mathcal{B}$ including Multi-source Knowledge Graph $\mathcal{G}$ and Multi-dimensional Interaction Repository $\mathcal{R}$, which unify educational data from diverse platforms and align semantically equivalent entities across datasets.
On top of this, a Multi-view Fusion Retrieval mechanism is applied to extract relevant contextual information for given inputs. 
The retrieved information is then formatted into a Structured Reliable-Context-based Prompt, allowing a frozen LLM to perform zero-shot prediction and generate interpretable reliable-context-based reports.
Together, these components enable RAG-KT to achieve robust and explainable KT.

\subsection{Multi-source Heterogeneous Knowledge Base Construction}
The goal of this stage is to construct a unified and extensible Multi-source Knowledge Graph $\mathcal{G}$ and Multi-dimensional Interaction Repository $\mathcal{R}$ that integrate data from multiple sources, so as to better model the relationships between students, questions, and knowledge concepts from different sources and serve as the foundation for subsequent multi-view retrieval.
Unlike existing heterogeneous knowledge graphs for KT, our $\mathcal{G}$ comprises six types of nodes: Knowledge Concepts ($K$), Question Groups ($QG$), Questions ($Q$), Students ($S$), Ability Levels ($A$), and Difficulty Levels ($D$). 
These nodes are connected via six corresponding types of edges, including $K-K$, $S-A$, $S-QG$, $QG-D$, $QG-K$, and $Q-QG$. 
Together, these components form a richly structured multi-relational graph. 

We first collect multiple publicly available KT datasets, then estimate students' ability levels and question difficulties using the 2-parameter logistic Item Response Theory (IRT-2PL) model:
\begin{equation}
    P_{s,q} = \frac{1}{1 + e^{-a_q(\theta_s - b_q)}}
\end{equation}
where $\theta_s$ denotes the ability of student $s$, $b_q$ is the difficulty of question $q$, and $a_q$ is the question’s discrimination parameter. 
The resulting $\theta_s$ and $b_q$ are normalized and categorized into three levels (e.g., Low, Medium, High) using a standard deviation-based thresholding rule:
\begin{equation}
    \text{Level}(x) = 
        \begin{cases}
            \text{Low}, & x \leq \mu - \sigma \\
            \text{Medium}, & \mu - \sigma < x < \mu + \sigma \\
            \text{High}, & x \geq \mu + \sigma
        \end{cases}
\end{equation}
To enhance the structural modeling of domain knowledge, we compile a collection of knowledge concepts and their Predecessor-Successors or Associative relationships from K-12 mathematics textbooks. This forms what we call the Basic KC Graph $\mathcal{G_{KC}}$, a subgraph composed of $K$ and $K-K$. 

Since $K$ representations vary across datasets, we design a KC matching pipeline to align dataset-specific terms with those in the $\mathcal{G_{KC}}$. 
For concept pairs with high semantic similarity, direct mapping is performed. For low-similarity cases, we leverage a LLM to review and align ambiguous concepts to the most semantically appropriate node in the $\mathcal{G_{KC}}$, thereby ensuring consistency and coverage across datasets. Detailed KC matching methods are provided in the Appendix~\ref{appendix:kc_match}.

Since the data for graph construction is multi-source, the resulting questions are diverse, which makes Q alignment necessary. 
Therefore, we introduce a unified intermediate node type $QG$, defined by a unique pairing of knowledge concept and difficulty level. Each $QG$ is constructed as:
\begin{equation}
    \text{QG}_{i,j} = \{ q \mid D_{q} = D_i, K_{q} = K_j\}
\end{equation}
This design allows questions from different datasets with the same semantic intent to be aligned, enabling downstream tasks to generalize across question sources.
In parallel with $\mathcal{G}$ construction, we build a structured $\mathcal{R}$ that stores student interaction history in the form of triplets $(S, QG, \text{Correctness})$, which are further enriched by other dimensions such as concept-level and difficulty-level performance. This $\mathcal{R}$ provides the contextual source for later retrieval.

\subsection{Multi-view Fusion Retrieval}
Retrieving high-quality and task-relevant context is crucial for enabling the LLM to perform accurate zero-shot prediction in KT.
Thus, this stage aims to retrieve high-quality, task-relevant contextual information from the $\mathcal{B}$, enabling zero-shot prediction by the LLM. 
Each prediction target consists of a target question $Q_{tgt}$ and the historical interaction sequence of the target student:
\begin{equation}
    S_{tgt} = \{(q_1, r_1), \dots, (q_n, r_n)\},  r_i \in \{0,1\}
\end{equation}
To support retrieval, we first identify the target knowledge concept $K_{tgt}$ associated with $Q_{tgt}$ via semantic matching over $\mathcal{G}$. 
To support fine-grained reasoning over different facets, we extract three task-specific subgraphs from $\mathcal{G}$, which correspond to different views. 
They capture relevant information from structural, behavioral, and ability perspectives to provide complementary views that enable more accurate and interpretable retrieval.
Each subgraph is paired with a corresponding similarity computation module to identify top-$k$ similar students $S_{sim}$ whose interactions are used to form the final prompt context.

\textbf{KC Interest Subgraph for Behavior-based Similarity.} 
The KC Interest Subgraph contains all nodes within a $n$-hop neighborhood of $K_{tgt}$, including related concepts $K_{rel}$, question group $QG_{tgt}$, and difficulty $D_{tgt}$ nodes. 
This subgraph defines the shared semantic scope for comparing student behaviors.
We construct a behavior vector $\vec{b}_s$ for each student $s$ over the subgraph nodes. For each node $n$, the behavior score is defined as:
\begin{equation}
    \text{Score}^n_s = [\alpha \cdot \text{Acc}_s^n + (1 - \alpha) \cdot \text{DWA}_s^n] \cdot \text{Conf}_s^n
\end{equation}
where $\alpha \in [0,1]$ is a configurable weight (default 0.5), $\text{Acc}_s^n$ is the historical accuracy, and $\text{DWA}_s^n$ is the Dynamic Weighted Accuracy \cite{dwa}:
\begin{equation}
    \text{DWA}_s^n = \frac{\sum_{i=1}^{N} \beta^{N - i} \cdot r_i}{\sum_{i=1}^{N} \beta^{N - i}}, \quad \beta \in (0,1)
\end{equation}
with $N$ attempts on node $n$ and $\beta$ controlling the recency decay (default 0.8). 
$\text{Conf}_s^n$ is the confidence score, we define the confidence score as the product of sample sufficiency and performance stability, where the former captures the quantity of rich and reliable context and the latter reflects the consistency of behavior. 
More attempts and more stable recent performance give higher confidence.
This design ensures that confidence increases only when a student has answered sufficiently many questions with stable performance.
The vector $\vec{b}_s$ is:
\begin{equation}
    \vec{b}_s = [\text{Score}_s^{n_1}, \text{Score}_s^{n_2}, \dots, \text{Score}_s^{n_T}]
\end{equation}
The behavior similarity is then computed using cosine similarity:
\begin{equation}
    \text{Sim}_{\text{bhv}} = 1 - \frac{\vec{b}_{tgt} \cdot \vec{b}_s}{\|\vec{b}_{tgt}\| \cdot \|\vec{b}_s\|}
\end{equation}

\textbf{Structural Path Subgraph for Structure-based Similarity.} 
This subgraph includes all $K$ nodes reachable from $K_{tgt}$ via $K-K$ edges in the domain graph $\mathcal{G_{KC}}$. 
It captures prerequisite and associative relationships around the target concept.
For each student $s$, we compute a structural path relevance score that reflects how closely the student's learned knowledge concepts are connected to the target concept $K_{tgt}$ within the domain graph $\mathcal{G}_{KC}$.
We define the student's structural path score as the average of the inverse path lengths:
\begin{equation}
    \text{Score}_{\text{struc}}^s = \frac{1}{N} \sum_{i=1}^{N} \frac{1}{\text{len}(K_{tgt}, K_i^s)}
\end{equation}
Where ${K_i^s}$ denotes the set of knowledge concepts that student $s$ has interacted with, and $N$ is the size of the set. $\text{len}(\cdot, \cdot)$ denotes the shortest path length between two nodes in $\mathcal{G}_{KC}$.
We normalize the absolute score difference into a similarity score:
\begin{equation}
    \text{Sim}_{\text{struc}} = 1 - \frac{|\text{Score}^P_{tgt} - \text{Score}^P_s|}{\text{Score}^P_{max} - \text{Score}^P_{min}}
\end{equation}
This reflects how well the student’s concept path aligns structurally with the target concept.

\textbf{Ability Subgraph for Ability-based Similarity.} 
The Ability Subgraph includes all $S$ and $A$ nodes and their $S-A$ links from the graph construction phase. 
Each student’s ability $\theta_s$ is estimated via IRT-2PL and normalized to $[0,1]$.
We define ability similarity as:
\begin{equation}
    \text{Sim}_{\text{abil}} = 1 - |\theta_{tgt} - \theta_s|
\end{equation}

\textbf{Fusion and Peer Retrieval.} 
The final similarity score is computed via weighted fusion:
\begin{equation}
\begin{aligned}
    \text{Sim}_\text{final} &= \lambda_1 \cdot \text{Sim}_{\text{bhv}} + \lambda_2 \cdot \text{Sim}_{\text{struc}} \\
                            &+ \lambda_3 \cdot \text{Sim}_{\text{abil}}
\end{aligned}
\end{equation}
with $\lambda_1 + \lambda_2 + \lambda_3 = 1$ as configurable hyperparameters. 
We retrieve the top-$k$ most similar students based on $\text{Sim}_{\text{final}}$, and extract their interaction records from $\mathcal{R}$.  
For each dimension (e.g., $K$, $D$, $QG$), we compute their aggregate performance as:
\begin{equation}
    \text{Perf}_s^{d} = (\text{Acc}_s^{d}, \text{DWA}_s^{d}, \text{Attempts}_s^{d}, \text{Conf}_s^{d})
\end{equation}
The same features are also computed for the target student, forming a structured, multi-perspective context $Ctx$ that is encoded into the LLM prompt for subsequent prediction and explanation.

\subsection{Explainable Analysis Report Generation}
To address the critical need for interpretable outcomes in educational practice, this stage guides the LLM to transform the retrieved context into a structured, human-readable report through a Structured Reliable-Context-based Prompt.
Considering that our method does not involve fine-tuning, prompts thus have a significant impact on the performance of LLMs, making it crucial to design a high-quality prompt.
A simple prompt example is shown in the right of Fig.~\ref{fig:fig1}, and the complete prompt is provided in the Appendix~\ref{appendix:prompt}.

\textbf{Input Structure.}  
The LLM prompt consists of four blocks: (1) target student and question metadata (ability $\theta_{tgt}$, concept $K_{tgt}$, difficulty $D_{tgt}$, etc.); (2) individual performance metrics across multiple dimensions; (3) peer and similar student aggregates obtained via multi-view fusion retrieval; (4) historical answer trajectory for trend analysis.

\textbf{Reasoning Framework.}  
The LLM is guided by a reasoning scaffold incorporating: (1) context reliability (e.g., attempt counts); (2) positive and negative attribution factors (e.g., peer success, unstable accuracy); (3) conflict resolution using structural and behavioral consistency; (4) calibration against peer datasets. This promotes transparent and consistent prediction generation.

\textbf{Output Format.}  
The report includes: (1) predicted probability and qualitative judgment; (2) student ability and knowledge mastery summary; (3) reliable-context-based explanation detailing decision rationale and risk analysis. This enables actionable feedback for personalized instruction. 

Given the diversity of actual teaching scenarios, qualitative analysis can better adapt to different educational contexts, capture nuanced learning states that quantitative data alone may miss, and make the feedback more intuitive and actionable for educators and students.

\section{Experiment}

\subsection{Experimental Configuration}

\begin{table*}[!t]
\caption{Main results on three benchmark datasets. Bold values denote the best performance; bold and underlined values indicate the second-best. $\text{RAG-KT}_\text{LLM}$ employs different LLMs as the final prediction model.}
\centering
\scalebox{0.85}{
    \begin{tabular}{@{}c|c|ccc|ccc|ccc@{}}
    \toprule
    \multirow{2}{*}{\textbf{Type}} & \multirow{2}{*}{\textbf{Baselines}} & \multicolumn{3}{c|}{\textbf{ASSIST09}} & \multicolumn{3}{c|}{\textbf{ASSIST12}} & \multicolumn{3}{c}{\textbf{DBE-KT22}} \\
     &  & ACC & AUC & F1 & ACC & AUC & F1 & ACC & AUC & F1 \\ \midrule
    \multirow{7}{*}{\textbf{\begin{tabular}[c]{@{}c@{}}DL-based\\ Methods\end{tabular}}} & DKT & 72.18 & 81.82 & 78.12 & 70.86 & 69.12 & 74.98 & 71.08 & 72.26 & 74.52 \\
     & DKT+ & 72.64 & 82.47 & 78.65 & 71.34 & 70.16 & 75.42 & 71.76 & 73.42 & 75.06 \\
     & AT-DKT & 72.62 & 82.96 & 79.16 & 71.58 & 71.28 & 75.84 & 72.32 & 74.18 & 75.48 \\
     & DKVMN & 71.32 & 81.28 & 77.92 & 70.51 & 68.73 & 74.76 & 71.47 & 72.03 & 74.65 \\
     & Deep-IRT & 71.96 & 80.98 & 77.84 & 71.02 & 69.19 & 75.03 & 72.04 & 72.81 & 75.09 \\
     & AKT & 73.76 & 84.11 & 80.28 & 73.09 & 72.79 & 77.82 & 73.61 & {\underline{\textbf{75.29}}} & 77.48 \\
     & GKT & 72.23 & 81.53 & 78.23 & 70.96 & 69.07 & 75.11 & 72.06 & 73.11 & 75.03 \\ \midrule
    \multirow{2}{*}{\textbf{\begin{tabular}[c]{@{}c@{}}LLM-based Methods \\ (Prompting)\end{tabular}}} & HISE-KT & 79.12 & {\underline{\textbf{84.31}}} & 82.03 & {\underline{\textbf{73.77}}} & {72.33} & 79.51 & 70.82 & 72.40 & 79.94 \\
     & EFKT & 64.85 & 61.10 & 73.59 & 63.03 & 66.79 & 66.83 & 63.50 & 68.28 & 66.44 \\ \midrule
    \multirow{4}{*}{\textbf{\begin{tabular}[c]{@{}c@{}}LLM-based Methods \\ (Fine-tuning)\end{tabular}}} & EPLF & 70.13 & 81.27 & 71.60 & 70.30 & 69.64 & 75.82 & 74.63 & 72.72 & 72.38 \\
     & LLM-KT & 78.68 & 83.55 & 81.03 & 72.23 & 72.57 & 78.73 & 77.48 & 75.27 & 78.25 \\
     & 2T-KT & \multicolumn{1}{l}{74.55} & \multicolumn{1}{l}{81.32} & \multicolumn{1}{l|}{80.65} & \multicolumn{1}{l}{72.80} & \multicolumn{1}{l}{71.60} & \multicolumn{1}{l|}{75.45} & \multicolumn{1}{l}{75.43} & \multicolumn{1}{l}{74.42} & \multicolumn{1}{l}{78.86} \\
     & CIKT & 75.17 & 82.27 & 80.27 & 72.63 & 71.89 & 77.67 & 76.38 & 74.73 & 77.14 \\ \midrule
    \multirow{3}{*}{\textbf{Ours}} & $\text{RAG-KT}_\text{GPT-4o}$ & 78.34 & 82.57 & 83.20 & 72.50 & 72.87 & 79.37 & 76.50 & 74.53 & 85.74 \\
     & $\text{RAG-KT}_\text{Qwen-Plus}$ & {\underline{\textbf{79.20}}} & 83.97 & {\underline{\textbf{85.06}}} & 72.60 & \underline{\textbf{73.35}} & {\underline{\textbf{80.68}}} & {\underline{\textbf{78.40}}} & 73.69 & {\underline{\textbf{86.60}}} \\
     & $\text{RAG-KT}_\text{DeepSeek-R1}$ & \textbf{80.00} & \textbf{85.74} & \textbf{85.75} & \textbf{74.80} & \textbf{73.89} & \textbf{82.60} & \textbf{78.89} & \textbf{76.32} & \textbf{86.80} \\ \bottomrule
    \end{tabular}
}
\label{tab:tab1}
\end{table*}
We evaluate RAG-KT on three public KT datasets: \textbf{ASSIST09} and \textbf{ASSIST12}, collected from the ASSISTments platform \cite{assessment}, and \textbf{DBE-KT22} \cite{dbe-kt22}, from an online course at the Australian National University. Additionally, we use the \textbf{Eedi} \cite{eedi} dataset from the NeurIPS 2020 Education Challenge to assess performance in a fully cold-start setting, where all students, questions, and platform are unseen. 

To ensure leakage-free evaluation, we adopt a student-level disjoint split. After segmenting each student’s history into sub-sequences of length 25 and using the last interaction in each sub-sequence as the prediction target, we randomly sample 1,000 test sequences and assign all interactions from the corresponding students to the test set; all remaining students are used for training. Hence, no student appears in both splits. Moreover, all reported results are averaged over five independent random splits/samplings.

We implement RAG-KT using three LLM backbones: GPT-4o, Qwen-Plus, and DeepSeek-R1 to evaluate generalizability across model families. The constructed knowledge graph includes 317 concepts, 593 prerequisite-successor relations, and 932 associative links, 34,171 students, 951 question groups, and over 3.3 million interaction records.
For the multi-dimensional retrieval stage, we set the weights of behavior, structure, and ability similarities to $\lambda_1 : \lambda_2 : \lambda_3 = 4 : 3 : 3$ (which was determined through an experiment, with the results shown in Table~\ref{appendix:tbl1}), emphasizing behavioral alignment while retaining structural and ability-aware reasoning. 
The KC Interest Subgraph uses a 2-hop neighborhood, and the maximum path length in the Structural Path Subgraph is capped at 10. 
Peer retrieval selects the top-2 most similar users per query. 
All models follow the same preprocessing and evaluation protocols for fair comparison.

\subsection{Baselines}
To comprehensively evaluate the effectiveness of RAG-KT, we compare our method against a diverse set of baselines, including traditional deep learning-based KT models and recent LLM-enhanced approaches. The baselines are grouped into the following categories:
\begin{itemize}
    \item \textbf{DL-based Methods}: DKT \cite{dkt}, DKT+ \cite{dkt+}, AT-DKT \cite{at-dkt}, DKVMN \cite{dkvmn}, Deep-IRT \cite{deep-irt}, AKT \cite{akt}, GKT \cite{gkt}.
    \item \textbf{LLM-based Methods (Prompting)}: HISE-KT \cite{hisekt}, EFKT \cite{efkt}, LOKT \cite{lokt}. 
    \item \textbf{LLM-based Methods (Fine-tuning)}: LLM-KT \cite{llm-kt}, EPLF \cite{neshaei}, CIKT \cite{cikt}, CLST \cite{clst}, 2T-KT \cite{2tkt}.
\end{itemize}
Specific descriptions of each baseline are provided in the Appendix~\ref{appendix:baselines}.

\begin{table*}[]
\caption{Ablation results on three benchmark datasets. Each row removes or alters a key component of the full RAG-KT framework to assess its impact.}
\centering
\scalebox{0.9}{
    \begin{tabular}{@{}l|ccc|ccc|ccc@{}}
    \toprule
    \multicolumn{1}{c|}{\multirow{2}{*}{\textbf{Methods}}} & \multicolumn{3}{c|}{\textbf{ASSIST09}} & \multicolumn{3}{c|}{\textbf{ASSIST12}} & \multicolumn{3}{c}{\textbf{DBE-KT22}} \\
    \multicolumn{1}{c|}{} & ACC & AUC & F1 & ACC & AUC & F1 & ACC & AUC & F1 \\ \midrule
    Full & \textbf{80.00} & \textbf{85.74} & \textbf{85.75} & \textbf{74.80} & \textbf{73.89} & \textbf{82.60} & \textbf{78.89} & \textbf{76.32} & \textbf{86.80} \\ \hline
    w/o Similar Students & 77.28 & 81.07 & 83.60 & 69.53 & 71.47 & 78.13 & 77.40 & 73.98 & 84.17 \\
    w/o Question Groups & 76.73 & 84.25 & 83.32 & 70.01 & 72.33 & 78.67 & 77.94 & 75.50 & 85.16 \\
    w/o KC Interest Subgraph & 77.46 & \textbf{\underline{85.49}} & \textbf{\underline{84.23}} & 70.01 & 73.06 & 78.54 & 77.40 & 75.23 & 85.08 \\
    w/o Structural Path Subgraph & 78.50 & 84.16 & 84.89 & 72.40 & 71.54 & 80.67 & 76.71 & \textbf{\underline{75.98}} & 85.32 \\
    Random Similar Students & \textbf{\underline{78.54}} & 83.62 & 84.03 & \textbf{\underline{73.36}} & \textbf{\underline{73.14}} & \textbf{\underline{81.56}} & \textbf{\underline{78.38}} & 75.72 & \textbf{\underline{86.43}} \\
    w/o Retrieval & 68.50 & 62.37 & 74.35 & 64.20 & 67.41 & 71.35 & 64.40 & 68.58 & 72.65 \\ \hline
    \end{tabular}
}
\label{tab:tab2}
\end{table*}

\subsection{Main Results}
Tab.~\ref{tab:tab1} presents the main evaluation results across three benchmark datasets. 
Overall, our proposed RAG-KT framework significantly outperforms all baseline models, including both traditional deep learning approaches and recent LLM-based KT methods. 
Specifically, $\text{RAG-KT}_\text{DeepSeek-R1}$ achieves the highest scores across all three datasets, demonstrating strong robustness and generalization. 
The second-best results are consistently achieved by $\text{RAG-KT}_\text{Qwen-Plus}$ in most cases, further validating the stability of our multi-view retrieval-enhanced design across different LLM backbones. 
Compared to traditional KT methods, which suffer from limited context-awareness and low adaptability in cross-domain settings, our framework provides substantial performance gains. 
Moreover, in contrast to LLM-based KT methods that rely solely on internal reasoning, our framework benefits from structured external grounding via graph and peer retrieval, yielding not only higher prediction accuracy but also enhanced interpretability. 
These results affirm the effectiveness of our retrieval-augmented framework in bridging structured modeling and LLM reasoning for KT.

\subsection{Ablation Results}

The ablation study results, as shown in the Tab.~\ref{tab:tab2}, provide clear evidence of the importance of each key component within the RAG-KT framework:

First, removing the Similar Students module (w/o Similar Students) leads to a substantial drop in performance across all datasets. This highlights the critical role of retrieving ability-aligned and structurally similar students for enhancing prediction.
Second, removing Question Group nodes (w/o Question Groups) also results in consistent performance degradation. As Question Groups serve as an abstraction that unifies knowledge concepts with difficulty levels across platforms, their absence weakens the model’s generalization and retrieval alignment capabilities.
Third, regarding retrieval subgraphs, both the KC Interest Subgraph and Structural Path Subgraph show notable impact when ablated.  This demonstrates the value of semantic neighborhood modeling and verifies the effectiveness of structured reasoning based on graph connectivity.

The Random Similar Students setting, which replaces the similarity with random selection, consistently underperforms compared to the full model. This demonstrates the necessity of our designed similarity functions that integrate structural, behavioral, and ability signals.
Lastly, removing the entire retrieval module (w/o Retrieval) leads to a sharp drop in performance, achieving results comparable to existing prompting methods. Experimental data confirms that this is the most critical component, fundamentally highlighting the value of extra context in KT.

In summary, all components of RAG-KT contribute positively to performance, with Question Groups and similarity-based retrieval playing especially vital roles in enhancing both prediction accuracy and interpretability.

\begin{figure}[!t]
    \centering
    \includegraphics[width=1\linewidth]{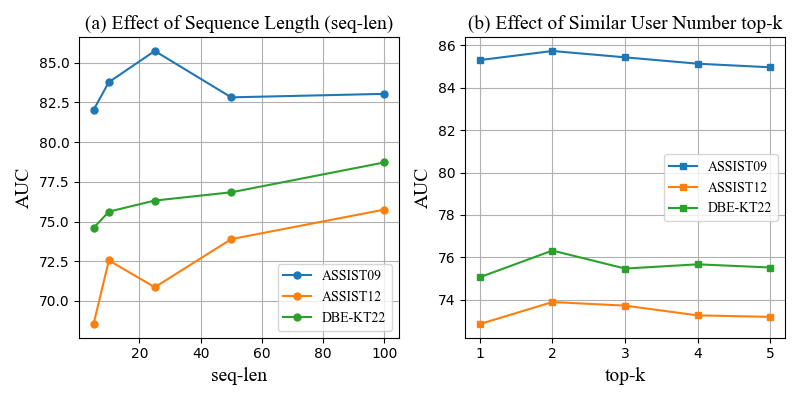}
    \caption{Parameter sensitivity analysis of RAG-KT on three datasets.
    }
    \label{fig:fig3}
\end{figure}
\subsection{Analysis about HyperParams}
Fig.~\ref{fig:fig3} presents the AUC performance of the proposed RAG-KT framework under different hyperparameter settings. 
In subplot (a), trends vary across datasets. ASSIST09 peaks at seq-len = 25, whereas ASSIST12 and DBE-KT22 generally improve as the sequence length increases, achieving their highest AUC at seq-len = 100. This indicates that while moderate context suits ASSIST09, the other datasets benefit significantly from longer historical sequences.
In subplot (b), we investigate the effect of the number of similar students (top-$k$) retrieved in the retrieval stage. Overall, performance remains stable when top-$k$ ranges from 2 to 4, with a slight peak at top-$k$ = 2, suggesting that including a small but relevant peer set provides the most beneficial calibration. Too many peers may introduce irrelevant or noisy patterns, slightly degrading performance. These results confirm that our framework is relatively robust to hyperparameter settings within reasonable ranges. Full results are provided in the Appendix~\ref{appendix:params}.

\subsection{Cold-start Results}
\begin{table}[]
    \caption{Cold-start Results}
    \centering
    \scalebox{0.7}{
        \begin{tabular}{@{}c|c|ccc@{}}
        \toprule
        \multirow{2}{*}{\textbf{Type}} & \multirow{2}{*}{\textbf{Baselines}} & \multicolumn{3}{c}{\textbf{Eedi}} \\
         &  & ACC & AUC & F1 \\ \midrule
        \multirow{4}{*}{\textbf{\begin{tabular}[c]{@{}c@{}}DL-based \\ Methods\end{tabular}}} & DKT & 49.54 & 50.23 & 49.61 \\
         & DKT+ & 49.98 & 50.62 & 49.84 \\
         & DKVMN & 51.58 & 51.20 & 52.64 \\
         & AKT & 53.55 & 49.69 & 47.68 \\ \midrule
        \multirow{2}{*}{\textbf{\begin{tabular}[c]{@{}c@{}}LLM-based Methods\\ (Cold-start)\end{tabular}}} & LOKT & \textbf{\underline65.88} & \textbf{\underline65.52} & \textbf{\underline57.90} \\
         & CLST & 61.45 & 63.60 & 42.23 \\ \midrule
        \textbf{Ours} & $\text{RAG-KT}_\text{DeepSeek-R1}$ & \textbf{68.00} & \textbf{74.40} & \textbf{70.43} \\ \bottomrule
        \end{tabular}
    }
    \label{tab:tab3}
\end{table}

As shown in Table~\ref{tab:tab3}, our proposed RAG-KT framework consistently outperforms both traditional deep learning models and recent LLM-based methods tailored for cold-start scenarios under the fully cold-start setting, where both students and questions are entirely unseen. In this setting, our constructed knowledge base and heterogeneous graph contain NO information from the Eedi dataset, and the only cross-platform signal comes from retrieval via the Question Group nodes.
This demonstrates the robustness and strong generalization ability of our method. By grounding predictions in a unified heterogeneous knowledge graph and leveraging multi-view fusion retrieval, our framework effectively mitigates challenges such as domain shift, data sparsity, and lack of prior interaction history. These issues severely limit the performance of other knowledge graph-based methods in cold-start environments.

\begin{figure}[!t]
    \centering
    \includegraphics[width=1\linewidth]{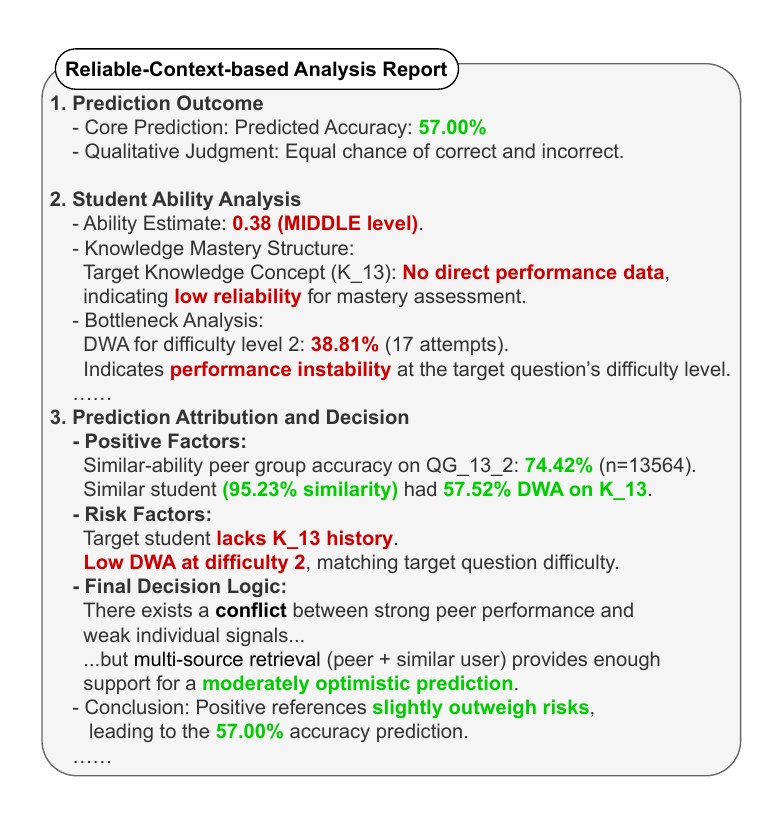}
    \caption{A case for our RAG-KT. The student answers this question correctly. The red font represents positive analysis and the green font represents negative analysis.
    }
    \label{fig:fig4}
\end{figure}

\subsection{Case Study}
We present a representative cold-start prediction case in Fig.~\ref{fig:fig4}. 
The model resolved the conflict between weak personal performance and strong external signals through a reasoning process that prioritized consistent multi-source rich context. This highlights the system’s context-aware decision mechanism, where structural alignment with peers and similar students enables accurate inference even in sparse data settings. A full example is provided in the Appendix~\ref{appendix:case}.

\subsection{Analysis about Report Quality}
\begin{table}[t]
    \caption{Scores for report quality evaluation on four dimensions: Explainability, Readability, Educational Usefulness, and Rigorousness}
    \centering
    \scalebox{0.7}{
        \begin{tabular}{@{}c|c|cccc|c@{}}
        \toprule
        \textbf{Baselines} & \textbf{Len.} & \textbf{Exp.} & \textbf{Read.} & \textbf{Edu.} & \textbf{Rig.} & \textbf{Total} \\ \midrule
        CIKT & 690 & 4.22 & \textbf{4.68} & \textbf{\underline{4.50}} & 3.83 & 17.23 \\
        EFKT & 82 & 2.70 & 2.69 & 2.40 & 2.26 & 10.05 \\ 
        HISE-KT & 350 & \textbf{\underline{4.72}} & 4.48 & 4.35 & \textbf{\underline{4.53}} & \textbf{\underline{18.08}} \\ \midrule
        $\text{RAG-KT}_\text{DeepSeek-R1}$ & 297 & \textbf{4.90} & \textbf{\underline{4.56}} & \textbf{4.88} & \textbf{4.90} & \textbf{19.24} \\ \bottomrule
        \end{tabular}
    }
    \label{tab:tab4}
\end{table}
To evaluate the quality of generated diagnostic reports, we adopt a mixed evaluation protocol that combines LLM-based scoring and expert human judgment. 
The detailed scoring mechanism is given in the Appendix~\ref{appendix:score}.
For automatic evaluation, we use DeepSeek-R1 \cite{deepseek} as an LLM-based evaluator to score generated reports.
For human evaluation, we invite three education experts to conduct an independent evaluation.
The raters are graduate students (M.S./Ph.D.) specializing in AI for Education, each with prior research experience. 
The final human evaluation scores are obtained by averaging across raters.
Due to the fact that pedagogical usefulness and report rigorousness are ultimately intended for human educators and learners, we report a final hybrid score that assigns a higher weight to human evaluation. Specifically, we combine the human and LLM-based scores with a 0.7:0.3 weighting scheme.

To ensure a comprehensive comparison, we selected three representative LLM-based baselines: EFKT as the pioneer of LLM-driven explainable KT, CIKT as the representative of fine-tuning  KT framework, and HISE-KT as the state-of-the-art method in interpretable KT.
As shown in Tab.~\ref{tab:tab4}, our proposed RAG-KT framework achieves the highest scores across most dimensions.
Although our framework achieves the best overall quality, CIKT slightly outperforms it in readability.
The reason for this is likely CIKT's core prompting methodology, which explicitly instructs the LLM to generate prose-style summaries organized by knowledge concepts. 
This naturally yields longer, more narrative outputs that provide detailed explanations, making them easier to read even if they are less data-dense. 
However, our framework strikes a better balance between brevity and informativeness across all dimensions.

\section{Discussion of Cost and Efficiency}
Despite its strong performance, RAG-KT introduces additional inference overhead compared with lightweight conventional KT models, mainly due to its retrieval-augmented pipeline and LLM-based reasoning stage. As shown in Table~\ref{tab:costtablt}, RAG-KT requires no additional training cost, while achieving the best AUC of 85.74 with a moderate inference latency of approximately 8 seconds per sample. Our profiling analysis shows that most of this time is spent on the final LLM inference stage, which takes around 6 seconds on average, while graph traversal and multi-view similarity computation contribute only a smaller portion of the overall latency. Compared with recent prompt-based explainable KT methods such as HISE-KT, which requires about 19 seconds per sample and achieves an AUC of 84.31, RAG-KT is substantially more efficient while also delivering better predictive performance. Although its latency remains higher than that of some traditional KT models, RAG-KT provides an important practical advantage beyond prediction accuracy: it additionally generates structured, evidence-grounded diagnostic reports, offering interpretability and actionable feedback for real educational use. This makes the framework especially suitable for scenarios such as after-class diagnosis, homework analysis, and teacher-facing learning support, where informative feedback is often more valuable than strict real-time response. In deployment, the inference cost can be further reduced by serving an open-source LLM locally or replacing the current backbone with a smaller model.
\begin{table}[!t]
\caption{Cost and Efficiency}
\centering
\scalebox{0.75}{
\begin{tabular}{@{}cccc@{}}
\toprule
\textbf{Methods} & \textbf{Trainging hours (h)} & \textbf{Inference latency (s)} & \textbf{AUC} \\ \midrule
EFKT             & /                         & 5                              & 61.10        \\
HISE-KT          & /                         & 19                             & 84.31        \\
CIKT             & 24                        & 3                              & 82.27        \\
EPLF             & 4                         & 1                              & 81.27        \\
LLM-KT           & 10                        & 3                              & 83.55        \\
2T-KT            & 20                        & 2.5                            & 82.27        \\ \midrule
RAG-KT           & /                         & 8                              & 85.74        \\ \bottomrule
\end{tabular}
}
    \label{tab:costtablt}
\end{table}

\section{Conclusion}
In this paper, we proposed RAG-KT, a retrieval-augmented knowledge tracing framework that integrated a multi-source heterogeneous knowledge graph with LLMs.
To enable cross-platform unification, RAG-KT introduced Question Group as an intermediate abstraction that aligned semantically and pedagogically similar questions across platforms into a shared representation.
By leveraging multi-view retrieval based on this unified space, RAG-KT achieved state-of-the-art performance and better interpretability on multiple datasets.
Notably, it maintained strong accuracy and interpretability even in fully cold-start scenarios, thereby demonstrating superior cross-platform generalization and real-world deployability.

\section*{Limitations}
Despite the promising performance and interpretability of RAG-KT, we acknowledge several limitations that require future investigation.
Compared to lightweight DL-based models, RAG-KT involves a multi-step process comprising graph retrieval and LLM generation, which incurs higher latency and API costs. 
In scenarios with extremely sparse data or poorly defined concept taxonomies where effective semantic alignment is impossible, the retrieval module may fail to extract meaningful context, potentially degrading performance.
Future work will explore distilling RAG-KT into smaller, local models to balance efficiency and performance.

\section*{Acknowledgments}
This research of Zhiyi Duan was funded by the National Natural Science Foundation of China (No. 62567005), and Natural Science Foundation of Inner Mongolia Autonomous Region of China (No. 2025MS06004).
This research of Rui Liu was funded by the General Program (No.62476146) of the National Natural Science Foundation of China, the Young Elite Scientists Sponsorship Program by CAST (2024QNRC001), the Outstanding Youth Project of Inner Mongolia Natural Science Foundation (2025JQ011), the Key R\&D and Achievement Transformation Program of Inner Mongolia Autonomous Region (2025YFHH0014), the Central Government Fund for Promoting Local Scientific and Technological Development (2025ZY0143).

\bibliography{custom}
 
\clearpage
\newpage

\appendix

\section{KC Matching Methodology}
\label{appendix:kc_match}
To address the challenge of varied terminologies for the same KC across different datasets (e.g., Pythagorean Theorem vs. Gougu Theorem), we implemented a pipeline to align all dataset-specific KCs ($KC^d$) with a canonical set of concepts in our pre-constructed $\mathcal{G_{KC}}$ ($KC^b$). 
To achieve this, we designed and employed a two-stage hybrid matching pipeline.

\textbf{Embedding-based Automatic Matching:} First, we utilize a embedding model to generate vector representations for all concepts in both $KC^b$ and $KC^d$. 
We then compute the cosine similarity for each ($KC^d_j$, $KC^b_i$) pair. 
If a pair exhibits the highest similarity score for a given $KC^d_j$ and this score exceeds a threshold of 0.85, we consider it a high-confidence match. 
This threshold is the best split point obtained by testing on a small validation set.

\textbf{LLM-based Semantic Alignment:} For pairs that fall below the threshold, or for more nuanced cases involving semantic equivalence despite lexical differences, we leverage a Large Language Model (LLM), such as GPT-4, for deep semantic alignment. We engineered a structured prompt that instructs the LLM to determine if two concepts are equivalent based on four strict criteria:
\begin{enumerate}
    \item Pedagogical Equivalence: Are the concepts pedagogically equivalent or do they have a very high degree of content overlap?
    \item Syllabus Coherence: Are they typically classified under the same specific module in an educational syllabus?
    \item Core Skill Identity: Do they teach the same fundamental mathematical essence or target the same core skills?
    \item Exclusion of Weak Relations: The relationship must be one of equivalence, not merely topical relevance or partial overlap.
\end{enumerate}

We explicitly evaluate alignment accuracy by first asking domain experts to manually align the KCs between the source and target platforms as ground truth. We then run our KC Match procedure and compare its outputs against this expert annotation. The resulting alignment accuracies are as shown  in Table~\ref{appendix:match}.
\begin{table}[h]
\caption{KC Match Accuracy.}
\centering
\begin{tabular}{@{}ccc@{}}
\toprule
\textbf{Source} & \textbf{Target} & \textbf{Accuracy} \\ \midrule
ASSIST09        & ASSIST12        & 96.33                       \\
ASSIST09        & DBE-KT22        & 92.67                       \\
ASSIST12        & DBE-KT22        & 96.50                       \\ \bottomrule
\end{tabular}
\label{appendix:match}
\end{table}

Through this carefully designed prompt, the LLM utilizes its extensive domain knowledge to accurately map the remaining ambiguous concepts. This two-stage hybrid approach ensures that our KC alignment process is both efficient for straightforward cases and accurate for complex semantic challenges.

\section{Structured Reliable-Context-based Prompt}
\label{appendix:prompt}
\begin{figure*}[]
    \centering
    \includegraphics[width=1\linewidth]{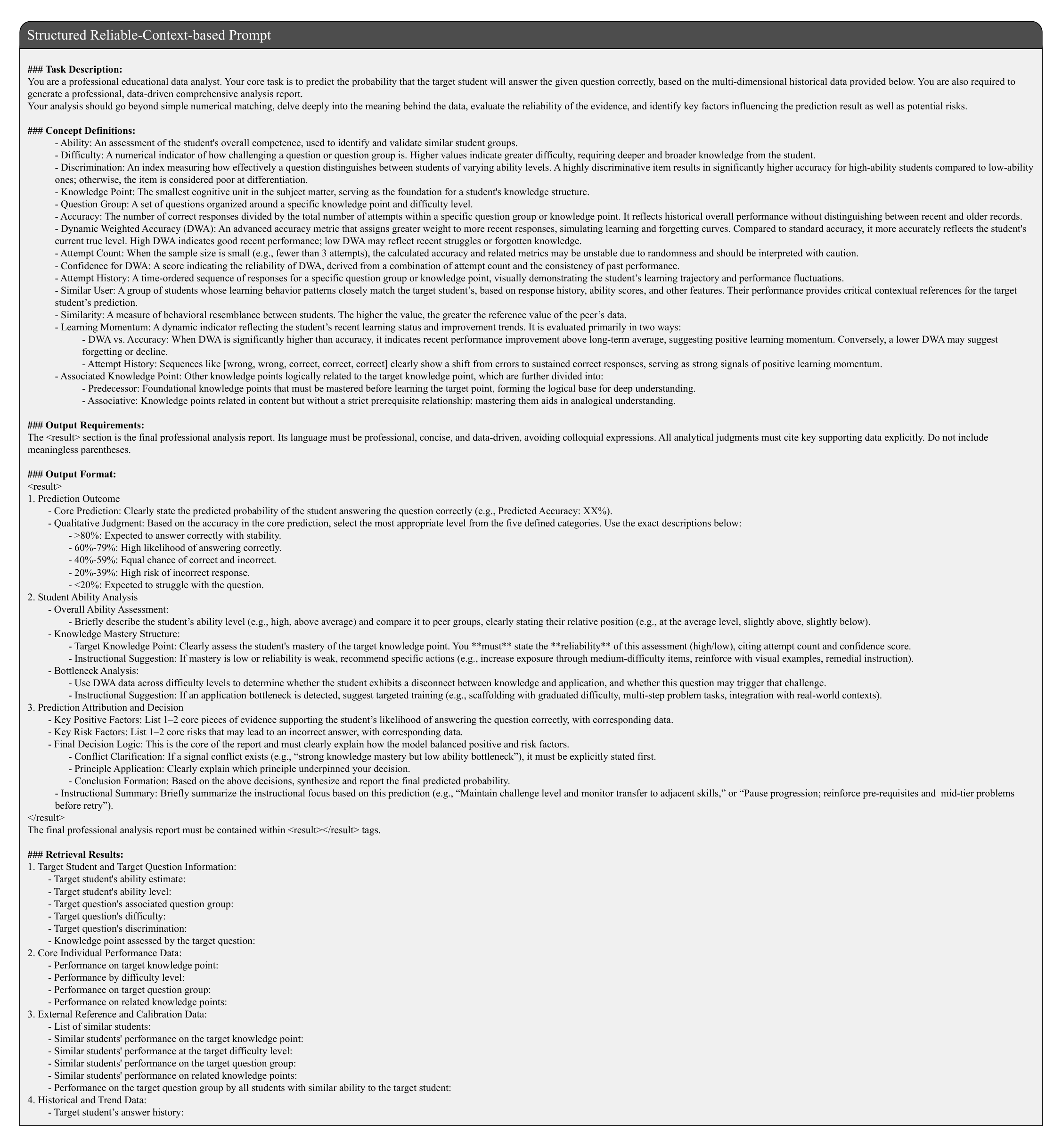}
    \caption{Structured Reliable-Context-based Prompt.}
    \label{appendix:fig1}
\end{figure*}

The complete Structured Reliable-Context-based Prompt is shown in Fig.~\ref{appendix:fig1}.

\section{Baselines}
\label{appendix:baselines}
Specific descriptions of each baseline:
\begin{itemize}
    \item \textbf{DL-based Methods}: These methods represent mainstream deep learning models for knowledge tracing: 
    \begin{itemize}
        \item \textbf{DKT}: A foundational RNN-based KT model that encodes student response sequences for prediction.
        \item \textbf{DKT+}: An extension of DKT that incorporates regularization and reconstruction losses to mitigate forgetting.
        \item \textbf{AT-DKT}: Integrates attention mechanisms to better capture dependencies in student sequences.
        \item \textbf{DKVMN}: A memory-augmented model that represents student knowledge states using key-value memory networks.
        \item \textbf{Deep-IRT}: Combines IRT principles with deep learning to improve modeling of student ability and item properties.
        \item \textbf{AKT}: Employs self-attention mechanisms and concept-aware modeling to capture temporal and contextual dynamics.
        \item \textbf{GKT}: Incorporates a graph structure to model skill dependencies and student progress.
    \end{itemize}
    
    \item \textbf{LLM-based Methods (Prompting)}: These methods employ LLMs in zero-shot or few-shot settings without parameter updates: 
    \begin{itemize}
        \item \textbf{EFKT}: Tracks knowledge states through few-shot prompting and generates natural language explanations.
        \item \textbf{LOKT}: Assigns differentiated weights to multiple-choice options to better capture students’ knowledge mastery, and serves as a comparison method for the fully cold-start setting in this study.
        \item \textbf{HISE-KT}: Synergizes heterogeneous information networks with LLMs, employing LLM-powered meta-path optimization and similar student retrieval to achieve accurate zero-shot prediction and reliable-context-based explanations.
    \end{itemize}

    \item \textbf{LLM-based Methods (Fine-tuning)}: These models involve explicit training or fine-tuning of LLMs for KT tasks: 
    \begin{itemize}
        \item \textbf{LLM-KT}: Proposes a plug-and-play prompting approach combining behavioral traces and textual context.
        \item \textbf{EPLF}: Evaluates LLMs’ zero-shot and fine-tuning ability to perform KT. We choose its fine-tuning setting for comparing.
        \item \textbf{CIKT}: Collaboratively fine-tunes two LLMs, a predictor and an analyst, to generate and use interpretable knowledge state descriptions.
        \item \textbf{CLST}: Reformulates the KT task as a natural language modeling problem using LLMs, and is used in this work to evaluate performance under fully cold-start scenarios.
        \item \textbf{2T-KT}: Leverages LLMs to simulate a teacher's thinking mode combined with knowledge graphs to address the new knowledge concept prediction problem.
    \end{itemize}

\end{itemize}

\section{HyperParams Results}
\label{appendix:params}
The complete HyperParams results are shown in Table~\ref{appendix:tbl1}.

\section{Case Study}
\label{appendix:case}
\begin{figure*}[!ht]
    \centering
    \includegraphics[width=1\linewidth]{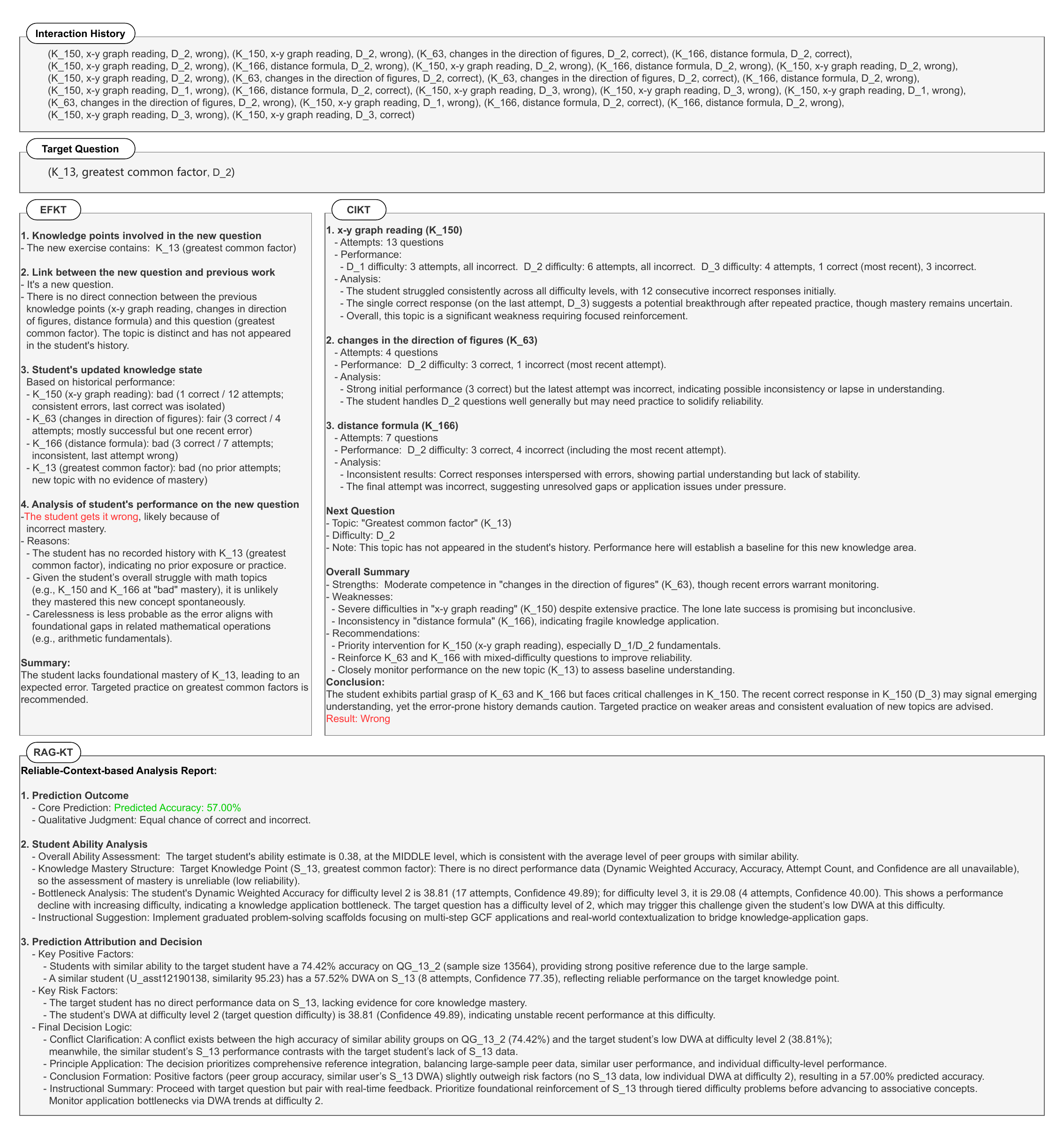}
    \caption{Case Study.}
    \label{appendix:fig2}
\end{figure*}

As shown in Fig.~\ref{appendix:fig2}, this section details a case study demonstrating RAG-KT compared with other models (EFKT and CIKT) in a cold-start scenario. The student answers this question correctly.

\section{Scoring Mechanism}
\label{appendix:score}
The quality of the generated analysis reports was evaluated based on four dimensions: Explainability, Readability, Educational Usefulness, and Rigorousness. Each dimension was scored on a 1 to 5 scale, with detailed criteria provided in the Table~\ref{appendix:tbl2}.


\begin{table*}[!ht]
\caption{Complete HyperParams results.}
\centering
\begin{tabular}{@{}c|c|ccc|ccc|ccc@{}}
\toprule
\multirow{2}{*}{\textbf{HyperParams}} & \multirow{2}{*}{\textbf{Value}} & \multicolumn{3}{c|}{\textbf{ASSIST09}} & \multicolumn{3}{c|}{\textbf{ASSIST12}} & \multicolumn{3}{c}{\textbf{DBE-KT22}} \\
 &  & ACC & AUC & F1 & ACC & AUC & F1 & ACC & AUC & F1 \\ \midrule
\multirow{5}{*}{\textbf{seq\_len}} & 5 & 76.77 & 82.03 & 83.62 & 69.34 & 68.54 & 78.52 & 76.68 & 74.59 & 84.58 \\
 & 10 & 78.48 & \textbf{\underline83.78} & 85.15 & 70.37 & 72.55 & 78.32 & 78.19 & 75.62 & 85.81 \\
 & 25 & \textbf{80.00} & \textbf{85.74} & \textbf{85.75} & 70.37 & 70.86 & 79.28 & \textbf{78.89} & 76.32 & \textbf{86.80} \\
 & 50 & 78.28 & 82.82 & 85.14 & \textbf{\underline74.80} & \textbf{\underline73.89} & \textbf{\underline82.6} & \textbf{\underline77.98} & \textbf{\underline76.84} & 85.59 \\
 & 100 & \textbf{\underline79.90} & 83.04 & \textbf{\underline86.04} & \textbf{75.83} & \textbf{75.75} & \textbf{82.76} & \textbf{78.89} & \textbf{78.72} & \textbf{\underline86.60} \\ \midrule
\multirow{5}{*}{\textbf{top\_k}} & 1 & \textbf{\underline79.79} & 85.31 & 85.55 & 73.79 & 72.85 & \textbf{\underline82.24} & 77.47 & 75.06 & 85.22 \\
 & 2 & \textbf{80.00} & \textbf{85.74} & \textbf{\underline85.75} & \textbf{74.80} & \textbf{73.89} & \textbf{82.60} & \textbf{78.89} & \textbf{76.32} & \textbf{86.80} \\
 & 3 & 79.51 & \textbf{\underline85.44} & 85.29 & \textbf{\underline74.20} & \textbf{\underline73.72} & 81.17 & \textbf{\underline78.51} & 75.47 & \textbf{\underline85.70} \\
 & 4 & 79.10 & 85.14 & 85.06 & 74.16 & 73.26 & 80.04 & 78.00 & \textbf{\underline75.67} & 85.33 \\
 & 5 & 78.80 & 84.97 & \textbf{85.82} & 73.18 & 73.19 & 80.88 & 77.65 & 75.52 & 85.25 \\ \midrule
\multicolumn{1}{l|}{\multirow{5}{*}{\textbf{$\lambda_1 : \lambda_2 : \lambda_3$}}} & \multicolumn{1}{l|}{1 : 1 : 1} & \multicolumn{1}{l}{79.21} & \multicolumn{1}{l}{\textbf{\underline{85.41}}} & \multicolumn{1}{l|}{85.13} & \multicolumn{1}{l}{73.95} & \multicolumn{1}{l}{73.52} & \multicolumn{1}{l|}{81.98} & \multicolumn{1}{l}{78.13} & \multicolumn{1}{l}{\textbf{\underline{76.15}}} & \multicolumn{1}{l}{86.24} \\
\multicolumn{1}{l|}{} & \multicolumn{1}{l|}{2 : 1 : 1} & \multicolumn{1}{l}{\textbf{\underline{79.72}}} & \multicolumn{1}{l}{85.30} & \multicolumn{1}{l|}{\textbf{\underline{85.45}}} & \multicolumn{1}{l}{\textbf{\underline{74.51}}} & \multicolumn{1}{l}{\textbf{\underline{73.66}}} & \multicolumn{1}{l|}{\textbf{\underline{82.31}}} & \multicolumn{1}{l}{\textbf{\underline{78.54}}} & \multicolumn{1}{l}{75.98} & \multicolumn{1}{l}{\textbf{\underline{86.59}}} \\
\multicolumn{1}{l|}{} & \multicolumn{1}{l|}{4 : 3 : 3} & \multicolumn{1}{l}{\textbf{80.00}} & \multicolumn{1}{l}{\textbf{85.74}} & \multicolumn{1}{l|}{\textbf{85.75}} & \multicolumn{1}{l}{\textbf{74.80}} & \multicolumn{1}{l}{\textbf{73.89}} & \multicolumn{1}{l|}{\textbf{82.60}} & \multicolumn{1}{l}{\textbf{78.89}} & \multicolumn{1}{l}{\textbf{76.32}} & \multicolumn{1}{l}{\textbf{86.80}} \\
\multicolumn{1}{l|}{} & \multicolumn{1}{l|}{3 : 4 : 3} & \multicolumn{1}{l}{79.58} & \multicolumn{1}{l}{85.15} & \multicolumn{1}{l|}{85.24} & \multicolumn{1}{l}{74.33} & \multicolumn{1}{l}{73.28} & \multicolumn{1}{l|}{82.07} & \multicolumn{1}{l}{78.21} & \multicolumn{1}{l}{75.74} & \multicolumn{1}{l}{86.22} \\
\multicolumn{1}{l|}{} & \multicolumn{1}{l|}{3 : 3 : 4} & \multicolumn{1}{l}{79.61} & \multicolumn{1}{l}{85.23} & \multicolumn{1}{l|}{85.39} & \multicolumn{1}{l}{74.45} & \multicolumn{1}{l}{73.41} & \multicolumn{1}{l|}{82.15} & \multicolumn{1}{l}{78.43} & \multicolumn{1}{l}{75.81} & \multicolumn{1}{l}{86.48} \\ \bottomrule
\end{tabular}
\label{appendix:tbl1}
\end{table*}

\begin{table*}[!ht]
    \caption{Scoring Mechanism.}
    \centering
    \begin{tabular}{@{}c|c|l@{}}
    \toprule
    \textbf{Dimension} & \textbf{Score} & \multicolumn{1}{c}{\textbf{Description}} \\ \midrule
    \multirow{5}{*}{\textbf{Explainability}} & 1 & Lacks any reasoning or presents a chaotic causal chain. \\
     & 2 & Vague causal logic that is difficult to understand. \\
     & 3 & Partially valid reasoning but lacks overall coherence. \\
     & 4 & The reasoning chain is largely complete and causal relationships are clear. \\
     & 5 & \begin{tabular}[c]{@{}l@{}}Fully reveals the reasoning process, accurately explaining why a prediction \\ was made.\end{tabular} \\ \midrule
    \multirow{5}{*}{\textbf{Readability}} & 1 & Chaotic, obscure, or incomprehensible language. \\
     & 2 & Verbose, with a disorganized and messy structure. \\
     & 3 & Largely clear language, but with logical leaps or excessive jargon. \\
     & 4 & Clear and well-structured expression with appropriate use of terminology. \\
     & 5 & \begin{tabular}[c]{@{}l@{}}Fluent, logically coherent, and concise language that is easy for the reader \\ to understand.\end{tabular} \\ \midrule
    \multirow{5}{*}{\textbf{Educational Usefulness}} & 1 & Offers no educational value and contains only generic statements. \\
     & 2 & Identifies issues too vaguely to provide guidance for students or teachers. \\
     & 3 & Points out specific problems and provides a preliminary analysis. \\
     & 4 & Clearly pinpoints a student's issues and offers insightful feedback. \\
     & 5 & \begin{tabular}[c]{@{}l@{}}Provides highly targeted and actionable suggestions that are significantly \\ helpful for teaching and learning.\end{tabular} \\ \midrule
    \multirow{5}{*}{\textbf{Rigorousness}} & 1 & Content is subjective, reasoning is arbitrary, and lacks any factual support. \\
     & 2 & Provides some context, but the argumentation is vague and unconvincing. \\
     & 3 & Generally context-based, but lacks detail or has a loose logical structure. \\
     & 4 & Supported by sufficient context, with meticulous logic and clear details. \\
     & 5 & \begin{tabular}[c]{@{}l@{}}Features a rigorous reasoning structure where all conclusions are explicitly \\ supported by clear context, with no logical fallacies.\end{tabular} \\ \bottomrule
    \end{tabular}
    \label{appendix:tbl2}
\end{table*}

\end{document}